\documentclass[conference, a4paper, finale, 10pt]{IEEEtran}
\usepackage[T1]{fontenc}
\usepackage{url}
\usepackage{cite}
\usepackage{booktabs}
\usepackage{bm}
\usepackage{amsmath, amssymb, amsfonts}
\usepackage{stmaryrd}
\usepackage{algorithmic}
\usepackage{graphicx}
\usepackage{textcomp}
\usepackage{xcolor}

\ifCLASSOPTIONcompsoc
\usepackage[caption=false, font=normalsize, labelfont=sf, textfont=sf]{subfig}
\else
\usepackage[caption=false, font=footnotesize]{subfig}
\fi

\newtheorem{proposition}{Proposition}

\def\BibTeX{{\rm B\kern-.05em{\sc i\kern-.025em b}\kern-.08em
T\kern-.1667em\lower.7ex\hbox{E}\kern-.125emX}}

\DeclareMathOperator{\inc}{\trianglelefteqslant}
\newcommand{\real}{\mathbb{R}}
\newcommand{\trans}{\top}
\newcommand{\ie}{i.e.\ }
\newcommand{\eg}{e.g.\ }

\begin{document}
    \title{Lying Graph Convolution: Learning to Lie for Node Classification Tasks}

    \author{\IEEEauthorblockN{Daniele Castellana}
    \IEEEauthorblockA{\textit{Department of Statistics, Informatics and Application} \\
    \textit{Università degli Studi di Firenze}\\
    Firenze, Italy \\
    daniele.castellana@unifi.it}}

    \maketitle

    \begin{abstract}
        In the context of machine learning for graphs, many researchers have empirically observed that Deep Graph Networks (DGNs) perform favourably on node classification tasks when the graph structure is homophilic (\ie adjacent nodes are similar).
        In this paper, we introduce Lying-GCN, a new DGN inspired by opinion dynamics that can adaptively work in both the heterophilic and the homophilic setting.
        At each layer, each agent (node) shares its own opinions (node embeddings) with its neighbours.
        Instead of sharing its opinion directly as in GCN, we introduce a mechanism which allows agents to lie.
        Such a mechanism is adaptive, thus the agents learn how and when to lie according to the task that should be solved.
        We provide a characterisation of our proposal in terms of dynamical systems, by studying the spectral property of the coefficient matrix of the system.
        While the steady state of the system collapses to zero, we believe the lying mechanism is still usable to solve node classification tasks.
        We empirically prove our belief on both synthetic and real-world datasets, by showing that the lying mechanism allows to increase the performances in the heterophilic setting without harming the results in the homophilic one.
    \end{abstract}

    \begin{IEEEkeywords}
        Deep Graph Networks, Opinion dynamics, Heterophilic graphs
    \end{IEEEkeywords}


    \section{Introduction}\label{sec:introduction}
    Graphs are natural abstractions that arise in many scientific fields such as chemistry~\cite{reiser_graph_2022}, physics~\cite{carleo_machine_2019}, and recommender systems~\cite{fan_graph_2019} to name a few.
    Due to this large number of applications, in recent years there has been an increasing interest in machine learning techniques that can process graph-structured data.
    Among them, Deep Graph Networks (DGNs)~\cite{bacciu_gentle_2020} are able to solve a task by generating node representations of the input graph.
    The generation process is based on neural~\cite{scarselli_graph_2009,micheli_neural_2009}, probabilistic~\cite{bacciu_probabilistic_2020}, or even untrained message passing~\cite{gilmer_neural_2017}.
    Researchers have empirically observed that such models perform favourably on node classification tasks when the graph structure is homophilic~\cite{mcpherson_birds_2001}, meaning that adjacent nodes in the graph share similar features or target labels to be predicted.
    In contrast, heterophilic graphs exhibit an opposite behaviour, and structure-agnostic baselines like a Multi-Layer Perceptron (MLP) proved to be better or very competitive compared to DGNs at classifying nodes under heterophily~\cite{zhu_beyond_2020,maekawa_beyond_2022}.

    The scope of this paper is to introduce Lying-GCN, a new DGN on Graph Convolution Network (GCN)~\cite{kipf_semi-supervised_2016} that is able to perform in both the homophilic and the heterophilic setting.
    The novelty of our proposal is a propagation schema which has a straightforward interpretation in terms of opinion dynamics, \ie the study of how preferences or opinions emerge and evolve in social networks.
    At each layer, we can interpret the node embeddings with $d$ channels as the private opinion of the agents (nodes) over a specific set of topics (channels).
    During the propagation, each agent listens to its neighbours to develop new opinions.
    While in GCN all the agents are truth-tellers since they share their private opinions with the neighbourhood directly, in Lying-GCN an agent can lie: it can share opinions that can be different from its private ones.
    In practice, when the agent $u$ communicates with the agent $v$, the lying mechanism generates a set of weights which are then multiplied element-wise with the opinion of $u$.
    Since the weights can be negative, the agent can lie by sharing an opinion that is opposite to its private one.
    To the extent of our knowledge, the only other work which also considers the opinion dynamics in the context of DGN is~\cite{bodnar_neural_2022}, where the authors introduce a new DGN denoted as Neural Sheaf Diffusion (NSD).
    Their proposal is grounded on the mathematical theory of \emph{cellular sheaves}, which has been already used to model the opinion polarisation~\cite{hansen_opinion_2021}.
    However, NSD cannot be interpreted in terms of agents and opinions.

    As a part of our work, we provide a characterisation of the lying mechanism in terms of dynamical systems.
    To this end, we derive the equivalent dynamical system defined as a set of ordinary different equations.
    The behaviour of such a system depends on the spectral property of the coefficient matrix, which in our case is the result of an element-wise multiplication between the symmetric normalised Laplacian of the graph and a matrix that contains the opinion weights.
    Our theoretical results show that, while the system asymptotically collapses to zero, the dynamics in the early stages are rich due to an oscillatory pattern caused by the presence of complex eigenvalues.
    Thus, we believe that the lying mechanism is still actionable to solve node classification tasks.

    We empirically test our hypothesis on both synthetic and real-world datasets, showing that 1) the lying mechanism allows to increase the performances of GCN in the heterophilic setting without hurting them in the homophilic ones, and 2) we are able to reach the performances of challenging baselines in the literature.

    The rest of the paper is organised as follows: in Section~\ref{sec:background} we introduce GCN and NSD, along with their interpretations in terms of opinion dynamics;
    in Section~\ref{sec:lying-gcn} we introduce our proposal, and we study the associated dynamical system;
    in Section~\ref{sec:other_related_works} we discuss other relevant approaches to increase the performances of DGN in the heterophilic setting;
    in Section~\ref{sec:exp_results} we describe our experiments, and we show the results obtained;
    finally, in Section~\ref{sec:conclusion}, we draw our conclusion.

    \section{Background} \label{sec:background}

    \subsection{Opinion Dynamics on Graphs}\label{subsec:opinion-dynamics-on-graphs}
    In the context of opinion dynamics, a graph $G = (\mathcal{V}, \mathcal{E})$ represents a social network where each node $v \in \mathcal{V}$ is an agent and each edge $(u,v) \in \mathcal{E}$ constitutes a pairwise (undirected) communication channel.
    For the sake of simplicity, we assume that the graph is connected and does not contain self-loops (\ie $(u,u) \notin \mathcal{E}$).
    The value $x_v \in \real$ represents the (binary) opinion of the agent $v$ about a topic: $x_v >0$ ($x_v<0$) indicates a positive (negative) opinion.
    All the opinions of all the agents can be concatenated into a single vector $\bm{x}\in \real^n$, where $n$ is the number of nodes in $G$.
    With abuse of notation, we use $\bm{x}_v$ to denote the opinion $x_v$ of the agent $v$.

    Since the agents can communicate with each other, the opinions evolve over time.
    A straightforward way to model the opinion dynamics is~\cite{Abelson64,taylor_1968_towards}:
    \begin{equation}
        \frac{d\bm{x}}{dt} = -\alpha L \bm{x}, \quad \alpha>0, \label{eq:graph_diffusion}
    \end{equation}
    where $L = D - A$ is the graph Laplacian; $A$ and $D$ are the adjacency matrix and the diagonal degree matrix, respectively.
    The Equation~\eqref{eq:graph_diffusion} is a system of Ordinary Differential Equations (ODEs) known as the \emph{graph heat diffusion} which results in asymptotically stable equilibrium where all the agents share the same opinion.
    This global fixed consensus does not represent the typical behaviour of opinion distributions in social networks;
    indeed, one of the most salient features of real-world opinions is the existence of polarisation or failure to come to a consensus, known as the community cleavage problem~\cite{friedkin2015problem}.

    To this end, the authors in~\cite{hansen_opinion_2021} propose to employ cellular sheaves theory~\cite{curry_sheaves_2014}.
    A cellular sheaf $\mathcal{F}$ is a mathematical object that augments a graph $G$ by specifying:
    \begin{itemize}
        \item a vector space $\mathcal{F}(v)$ for each vertex $v \in \mathcal{V}$;
        \item a vector space $\mathcal{F}(e)$ for each edge $e \in \mathcal{E}$;
        \item a linear map $\mathcal{F}_{v \inc e}: \mathcal{F}(v) \rightarrow \mathcal{F}(e)$ for each incident vertex-edge pair $v \inc e$.
    \end{itemize}
    The vector spaces $\mathcal{F}(v)$ are called the \emph{stalks} over $v$, and the linear maps $\mathcal{F}_{v \inc e}$ are the \emph{restriction maps}.

    Through the lens of the opinion dynamics, the stalk $\mathcal{F}(v)$ represents the \emph{opinion space} of agent $v$; thus, the opinion of $v$ can be any element in $\mathcal{F}(v)$, \ie $x_v \in \mathcal{F}(v)$.
    Rather than sharing directly their opinions, the agents generate a discourse based on them.
    The generation is performed by the restriction maps: given an edge $e=(u,v)$, the discourse of $u$ and $v$ is given by $\mathcal{F}_{u \inc e}(x_u)$ and $\mathcal{F}_{v \inc e}(x_v)$, respectively.
    Both discourses lie in the same space: the edge stalk $\mathcal{F}(e)$; hence, the edge stalks are denoted as \emph{discourse spaces}.

    The cellular sheaf modifies also the interpretation of consensus.
    Two agents $u$ and $v$ linked by the edge $e$ reach a consensus when they agree in discourse space, \ie $\mathcal{F}_{u \inc e}(x_u) = \mathcal{F}_{v \inc e}(x_v)$.
    Note that this does not imply that $u$ and $v$ have the same opinions: it means that their expressions of personally held opinions have the appearance of agreement.
    The opinion dynamics on cellular sheaf which converge to a global edge consensus is:
    \begin{equation}
        \frac{d\bm{x}}{dt} = -\alpha L_\mathcal{F} \bm{x}, \quad \alpha>0. \label{eq:graph_sheaf_diffusion}
    \end{equation}

    The vector $\bm{x}$ is a \emph{0-cochain} and it records all the agent opinions;
    thus, it lies on the space $C^0(G; \mathcal{F})$ obtained by merging all the node stalks, \ie $C^0(G; \mathcal{F}) = \bigoplus_{v \in \mathcal{V}} \mathcal{F}(v)$ (where $\oplus$ denotes the direct sum of vector spaces). $L_\mathcal{F}: C^0(G; \mathcal{F}) \rightarrow C^0(G; \mathcal{F})$ is a linear operator denoted as \emph{sheaf Laplacian} and it is defined as $L_\mathcal{F}(\bm{x}_v) = \sum_{e=(u,v)} \mathcal{F}_{v \inc e}^\trans(\mathcal{F}_{v \inc e}\bm{x}_v-\mathcal{F}_{u \inc e}\bm{x}_u)$.

    For the purpose of our work, we can assume that all the stalks are vector spaces of size $d$, \ie $\mathcal{F}(v) = \mathcal{F}(e) = \real^d$.
    Thus, the opinion $x_v \in \real^d$ records the intensities of opinions or preferences of agent $v$ on $d$ different topics.
    Hence, a 0-cochain $\bm{x}$ is a vector of size $nd$, the restriction maps $\mathcal{F}_{v \inc e}$ are matrices $d \times d$, and the sheaf Laplacian $L_{\mathcal{F}}$ is a symmetric matrix $nd \times nd$.
    Finally, it is worth pointing out that the graph heat diffusion equation is a cellular sheaf with $d=1$ and all the restriction maps equal to $1$.





    \subsection{Diffusion \& Deep Graph Networks}\label{subsec:diffusion-&-deep-graph-networks}
    DGNs aim to learn a node representation which solves a given task.
    The Graph Convolutional Network (GCN)~\cite{kipf_semi-supervised_2016} is a DGN which computes the node embeddings with the following layer-wise propagation rule:
    \begin{equation}
        \bm{h}'_u = \sigma \left( W\left(\tilde{S}_{uu}\bm{h}_u + \sum_{v \in \mathcal{N}_u} \tilde{S}_{uv}\bm{h}_v\right)\right), \label{eq:gcn_equation}
    \end{equation}
    where $\bm{h}_u \in \real^{d}$ and $\bm{h}'_u \in \real^{d'}$ are the input and output node embeddings of the layer, respectively; $\sigma$ is a non-linear activation function, while $W \in \real^{d' \times d}$ is a weight matrix that is adjusted during the training.
    The matrix $\tilde{S}=\tilde{D}^{-\frac{1}{2}}\tilde{A}\tilde{D}^{-\frac{1}{2}}$ is the normalise adjacency matrix, $\tilde{A} = A + I$ and $\tilde{D}=D +I$ are the augmented adjacency and degree matrix, respectively.
    The same equation can be written in matrix form:
    \begin{equation}
        H' = \sigma\left(\tilde{S}HW^\trans\right)= \sigma\left(\left(I - \tilde{L}^{sym}\right)HW^\trans\right), \label{eq:gcn_equation_matrix}
    \end{equation}
    where $\tilde{L}^{sym}=I-\tilde{S}$ is the augmented symmetrically normalised Laplacian, and $H \in \real^{n\times d}$ is the matrix obtained by stacking all the hidden representations $\bm{h}_u$ for all the nodes $u\in \mathcal{V}$.
    This equation makes clear the relation between GCN and the graph heat diffusion in Equation~\eqref{eq:graph_diffusion}: in fact, the GCN computation can be interpreted as a finite difference approximation of the graph heat diffusion process (setting apart the weight matrix $W$ and the non-linearity $\sigma$)~\cite{wang_dissecting_2021}.
    The graph Laplacian is applied independently to each column of $H$, thus the hidden representation $\bm{h}_u$ represents the opinions of $u$ over $d$ topics which evolves separately.
    When the agents reach a consensus, it means that the hidden node representations are similar.
    In the DGN context, this behaviour is denoted as \emph{oversmoothing}~\cite{li_deeper_2018,chen_measuring_2020}.

    Neural Sheaf Diffusion (NSD)~\cite{bodnar_neural_2022} is a recent DGN which takes advantage of the sheaf theory to overcome the GCN limitations.
    The layer-wise propagation of the node embeddings in NSD is given by the following rule:
    \begin{equation}
        H' = H - \sigma\left(L_{\mathcal{F}} \left(I \otimes V\right) H W\right), \label{eq:neaural_sheaf_diffusion}
    \end{equation}
    where $H \in \real^{nd \times f}$ and  $H' \in \real^{nd \times f'}$ are the input and the output node embeddings. $L_{\mathcal{F}} \in \real^{nd \times nd}$ is the sheaf Laplacian, $V \in \real^{d \times d}$ and $W \in \real^{f \times f'}$ are the weight matrices.
    Again, $\sigma$ is a non-linear activation function.
    NSD is closely related to the cellular sheaf diffusion process in Equation~\eqref{eq:graph_sheaf_diffusion} since NSD propagation schema is equivalent to a finite difference approximation of it (setting apart the weight matrices and the non-linearity).

    A key point in NSD is the definition of the sheaf Laplacian $L_{\mathcal{F}_l}$.
    While in GCN the Laplacian is completely defined by the graph structure, in NSD the sheaf Laplacian depends on the restriction maps $\mathcal{F}_{v\inc e}$ of the cellular sheaf which must be defined.
    The authors in~\cite{bodnar_neural_2022} show that, given a node classification task on a graph $G$, it exists a cellular sheaf on $G$ such that its stable equilibria can linearly separate the node classes of $G$.
    Unfortunately, such a cellular sheaf depends on the node class labels which are not known at training time: thus, it can be defined only by an oracle.
    Notably, the sheaf Laplacian $L_{\mathcal{F}_l}$ is always symmetric.

    At each layer, NSD tries to learn the right sheaf using locally available information.
    Let $e=(u,v)$ an edge in $G$, NSD approximate the restriction maps $\mathcal{F}_{v\inc e}$ by employing a parametric function approximator, \ie $\mathcal{F}_{v\inc e}= \sigma\left(Q\left[\bm{h}_v\parallel \bm{h}_u\right]\right)$ where the symbol $\parallel$ denotes the concatenation of two vectors.
    Three different versions of NSD have been proposed: Diag-NSD, O(d)-NSD and Gen-NSD which impose a diagonal, orthogonal and no constraints on the restriction maps, respectively.
    While NSD stems from cellular sheaves on graphs, there is no direct interpretation of NSD as opinion diffusion.
    The hidden node representations are no longer vectors of size $d$ (as it happens in the GCN): instead, they are matrices of size $d \times f$.
    This worsens also the computational complexity since it scales w.r.t. $d^2\times f^2$; in practice, only small values of $d$ have been employed (from 1 to 5).

    \section{Lying Graph Convolutional Network} \label{sec:lying-gcn}
    We propose Lying-GCN, a new DGN architecture which computes (at each layer) the node embedding as:
    \begin{align}
        \bm{z}_{v\rightarrow u} &= \tanh\left(V[\bm{h}_v \parallel \bm{h}_u]\right) \label{eq:z_building}\\
        \bm{m}_{v\rightarrow u} &= \bm{z}_{v\rightarrow u} \odot \bm{h}_v \label{eq:m_building}\\
        \bm{h}'_u &= \sigma\left(W\left(\tilde{S}_{uu}\bm{h}_u + \sum_{v\in \mathcal{N}_u}\tilde{S}_{uv}\bm{m}_{v\rightarrow u} \right)\right) \label{eq:lyingGCN_prop},
    \end{align}
    where $\odot$ is the element-wise multiplication, $\bm{z}_{v\rightarrow u}$ and $\bm{m}_{v\rightarrow u}$ are vectors of size $d$, and the matrices $V \in \real^{2d \times d}$ and $W \in \real^{d'\times d}$ are the parameters of the layer.

    The Lying-GCN propagation has a straightforward interpretation in terms of opinions.
    At each layer, each agent (node) $u$ has a private opinion $\bm{h}^l_u \in \real^d$ over $d$ topics.
    The initial opinion $\bm{h}^0_u$ can be obtained as a linear combination of the input feature $\bm{x}_u \in \real^f$, \ie $\bm{h}_u^0 = W_0 \bm{x}_u$.
    The agents do not share directly their opinions;
    as in the cellular sheaf diffusion, they can talk to each other over the communication links (edges) by making a discourse.
    Let $e=(u,v)$ an edge,  agent $v$ produce a discourse $\bm{m}_{v\rightarrow u} \in \real^d$ for agent $u$ starting from its opinions $\bm{h}_v$.
    Hence, the agent can lie to its neighbour by multiplying its opinion with a weight $\bm{z}_{v \rightarrow u} \in \real^d$.
    The weight has a value $z_d \in [-1,1]$ for each topic $d$; if $z_d=1$, agent $v$ is not lying and it will share its real opinion on $d$ to $u$.
    Conversely, if $z_d=-1$, agent $v$ is lying and it will share the opposite of its opinion on $d$.

    After the propagation, a weight matrix $W$ and a non-linearity $\sigma$ are applied node-wise to generate the new opinions (node embeddings).
    This step can be interpreted as a reflection period, where each agent reasons about its new information to build new opinions over a (possibly different) set of topics $\{1,\dots, d'\}$.
    Since the weight matrix is shared among all nodes, all the agents share the same way of thinking.

    By comparing Equation~\eqref{eq:gcn_equation} and Equation~\eqref{eq:lyingGCN_prop}, we can recover the GCN propagation by setting all the opinion weights $\bm{z}$ equals to one.
    In this way, the message $\bm{m}_{v \rightarrow u}$ sent from $v$ to $u$ is equal to its hidden state $\bm{h}_v$. 
    Lying-GCN also shares some similarities with NSD, since the message construction in Equation~\eqref{eq:m_building} can be interpreted as the application of a diagonal transportation maps such that $\text{diag}\left(\mathcal{F}_{u \inc e}^\trans \mathcal{F}_{v \inc e}\right) = \bm{z}_{v\rightarrow u}$, where $\text{diag}(B)$ is the vector formed by the elements on the main diagonal of the matrix $B$.
    In our proposal, the diffusion process is not symmetric since $\bm{z}_{v\rightarrow u} \neq \bm{z}_{u\rightarrow v}$.
    This is not true in Diag-NSD, where it holds $\left(\mathcal{F}_{u \inc e}^\trans \mathcal{F}_{v \inc e}\right) = \left(\mathcal{F}_{v \inc e}^\trans \mathcal{F}_{u \inc e}\right)^\trans$.

    \subsection{Computational Complexity}\label{subsec:computational-complexity}
    The computation of the messages requires $O(md^2 + md) = O(md^2)$ operations, where $m$ is the number of edges in the graph.
    The node embedding propagation has the same complexity of GCN, \ie $O(nd^2 + md)$.
    Thus, the final complexity of a Lying-GCN layer is $O(nd^2 + md^2)$.
    Note that, compared with NSD, the computational complexity depends only on the number of topics $d$.

    \subsection{The Lying Diffusion Process}\label{subsec:the-lying-diffusion-process}

    \begin{figure*}
        \centering
        \subfloat{
            \includegraphics[width=0.25\textwidth]{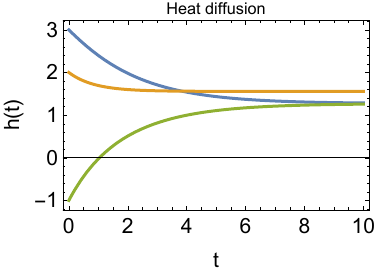}
            \label{fig:heat_ODE_solution}
        }
        \subfloat{
            \includegraphics[width=0.25\textwidth]{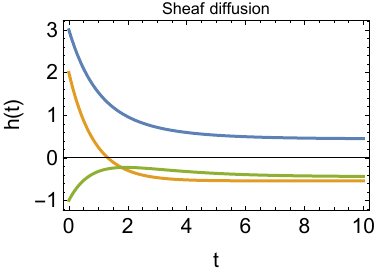}
            \label{fig:sheaf_ODE_solution}
        }
        \subfloat{
            \includegraphics[width=0.3\textwidth]{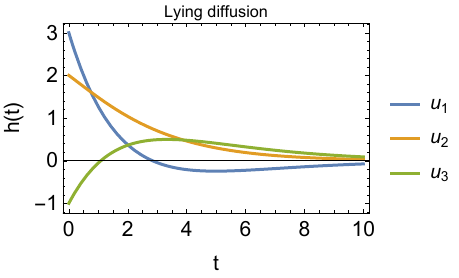}
            \label{fig:lying_ODE_solution}
        }
        \caption{From left to right, the evolution of the heat, sheaf and lying diffusion process on a chain graph with three nodes. In each plot, we show how the value attached to each node evolves over time.}
        \label{fig:ODE_solutions}
    \end{figure*}

    To better understand the effect of the opinion weights, it is convenient to rewrite the propagation schema in a matrix form.
    For the sake of simplicity, we focus on a single topic and we ignore both the non-linearity $\sigma$ and the weight matrix $W$.
    Thus, let $\bm{h} \in \real^n$ a vector which contains all the opinion $h_u \in \real$ of the agents, we can rewrite the propagation as:
    \begin{equation}
        \bm{h}' = \tilde{S} \odot \left(Z+I\right)\bm{h},\label{eq:equation}
    \end{equation}
    where $Z \in \real^{n \times n}$ is a non-symmetric matrix which contains the opinion weight, \ie $Z_{uv} = \tanh\left(Q[h_v \parallel h_u]\right)$ if $(u,v) \in \mathcal{E}$ and $Z_{uv:} = 0$ otherwise.
    The identity $I$ allows to preserve the self-loops added in $\tilde{S}$.
    By recalling the definition of the Laplacian $\tilde{L}^{sym}=I-\tilde{S}$, we obtain:
    \begin{equation}
        \begin{aligned}
            \bm{h}' &= \left((I-\tilde{L}^{sym}) \odot \left(Z+I\right)\right)\bm{h} \\
            &=\left(I-\tilde{L}^{sym}\odot \left(Z+I\right)\right)\bm{h},
        \end{aligned}\label{eq:equation2}
    \end{equation}
    where the last equality holds since we assume that there are no self-loops in the input graph (\ie $Z_{uu}=0$ for each node $u\in\mathcal{V}$, which implies $I\odot Z = 0$).
    If we assume that the matrix $Z$ is the same for each layer, the Lying-GCN propagation schema is a numerical discretisation of the following system of ODEs:
    \begin{equation}
        \frac{d\bm{h}}{dt} = - \left(\tilde{L}^{sym}\odot \left(Z+I\right)\right) \bm{h}, \label{eq:lying_diffusion}
    \end{equation}
    whose solution is:
    \begin{equation}
        \bm{h}(t) = e^{-Et} \bm{h}(0) = \sum_{i=i}^n c_ie^{-\lambda_i t}\bm{u}_i,\label{eq:equation3}
    \end{equation}
    where $E= (\tilde{L}^{sym}\odot \left(Z+I\right))$, and $(\lambda_i$, $\bm{u}_i)$ its $i$-th eigenvalue- eigenvector pair; $\{c_1, \dots c_n\}$ are constants that depend on the initial state $\bm{h}(0)$.
    Thus, the behaviour of the system depends on the spectral properties of $E$.

    \begin{proposition}
        The real part of each non-zero eigenvalue of $E$ is strictly positive.
        \label{prop:real_eigenvalue_E}
    \end{proposition}

    Proposition~\ref{prop:real_eigenvalue_E} (the proof is in Appendix~\ref{sec:appendix}) ensures us that the lying diffusion process converges to the zero vector since $\lim_{t\rightarrow +\infty} e^{-\lambda_it}=0$ for all $\lambda_i>0$.
    However, since $E$ is no longer symmetric, we have no guarantee that its eigenvalues are real values.
    The presence of complex eigenvalues enriches the process by adding oscillating dynamics.
    The oscillations derive from the exponentiation of the eigenvalues imaginary part since $e^{i\theta} = \cos \theta + i \sin \theta$ due to Euler's formula.

    In Figure~\ref{fig:ODE_solutions} we show the behaviour of three different examples of diffusion processes on a very simple chain graph $u_1 \leftrightarrow u_2 \leftrightarrow u_3$.
    The leftmost plot depicts the evolution of the graph heat diffusion process.
    As expected, the process converges to a state which depends on the node degree and the initial condition~\cite{li_deeper_2018}.
    Note that all the nodes have the same opinion since all the values are positive.
    The plot in the middle represents the evolution of a sheaf Laplacian with $d=1$ and edge weights $-1$ and $1$ on the edges $u_1 \leftrightarrow u_2$ and $u_2 \leftrightarrow u_3$, respectively.
    Thanks to the negative edge weight, the opinions of $u_1$ and $u_2$ diverge (regardless of the initial condition).
    In the rightmost plot, we show the evolution of our proposal: the lying diffusion process.
    In this case, the only negative weight is on the edge $u_2 \rightarrow u_1$, \ie the agent $u_2$ lies to $u_1$, but not vice versa.
    As expected, the process converges to 0.
    However, the process has more variability due to the complex part of the eigenvalues: for example, it is the only process where it exists an instant $t_i$ such that $u_i(t_i) > u_{j\neq i}(t_i)$ for all $i\in\{1,2,3\}$.
    Finally, it is worth highlighting that by carefully choosing the values in $Z$, the lying diffusion process can mimic both the heat and the sheaf diffusion.

    The analysis conducted above suggests that Lying-GCN may suffer from oversmoothing.
    According to~\cite{yan2022two, bodnar_neural_2022}, the oversmoothing phenomena is related to the poor performances in the heterophilic setting;
    thus, it could be reasonable to ask if our proposal is still actionable to solve node classification tasks.
    We believe that the answer is positive since, in practice, Lying-GCN can learn a different opinion weights matrix $Z$ at each layer.
    Thus, it can adaptively change the behaviour of the propagation schema to improve the performance.
    We empirically prove our belief in Section~\ref{sec:exp_results}.

    \section{Other Related Works} \label{sec:other_related_works}
    As in our proposal, Graph Attention Network (GAT)~\cite{velickovic_graph_2018} and Graph Neural Networks with Feature-wise Linear Modulation (GNN-FiLM)~\cite{brockschmidt2020film} define a learnable mechanism to build a message from node $u$ to its neighbour $v$ during the propagation step.
    For example, GAT learns an attention coefficient $\alpha_{vu}$ which measures the importance of node $u$´s embeddings to determine the new embeddings of $v$.
    While both models can learn asymmetric weights, they cannot model negative interaction among nodes since the weights are always positive.
    The inability to model negative interactions worsens the performance in the heterophilic setting.

    Other DGN architectures can learn negative values along edges.
    For example, Frequency Adaptation Graph Convolutional Networks (FAGCN)~\cite{bo2021beyond} employs an attention mechanism to learn the proportion of low-frequency and high-frequency signals to consider during propagation.
    Interestingly, this can be reduced to learning a weight $\alpha_{vu} \in [-1,1]$ over the edge $(u,v)$.
    A similar approach is employed in GGCN~\cite{yan2022two}, where the authors employ also a correction term on edges to compensate for the different behaviour between low-degree and high-degree nodes.
    However, in both approaches, the (possibly negative) edge weights are shared among all the channels of the node embeddings.
    On the contrary, our proposal can learn a different weight to each topic by allowing the definition of a different diffusion process for each topic.

    Finally, other models in the literature (\eg $\text{H}_2\text{GCN}$\cite{zhu2020beyond}, FSGNN~\cite{maurya2022simplifying}) address the heterophilic setting without modifying the propagation schema.
    Instead, they use techniques such as defining a different set of weights for ego- and neighbour-embeddings, considering $k$-hop neighbours, and concatenating all the intermediate node-representation.
    It is worth highlighting that these approaches are orthogonal to our proposal and they could be considered also in Lying-GCN\@.

    \section{Experiments \& Results} \label{sec:exp_results}
    In this work, we focus on node classification tasks where the goal is to assign a label to all the nodes in the input graph $G=(\mathcal{V}, \mathcal{E})$.
    For each node $u \in \mathcal{V}$, we denote the input features as a vector $\bm{x}_v \in \real^f$, and the output label as $y_u \in \mathcal{C}$ (where $\mathcal{C}$ is the set of all possible labels).
    All the models are equipped with an input layer and a linear classifier to process the input features and generate the output labels, respectively.

    In our experiments, we focus on two DGNs obtained by applying the lying message propagation schema: Lying-GCN and Lying-GCNII. While the former has been already discussed in Section~\ref{sec:lying-gcn}, the latter is obtained by combining the lying propagation schema with GCNII~\cite{chen_simple_2020}.
    GCNII is a deep GCN model that resolves the oversmoothing problem by employing (at each layer) a skip connection from the input feature (possibly pre-processed), and by adding an identity matrix to the weight matrix.
    The code of our experiments is publicly available \footnote{\url{https://github.com/danielecastellana22/lying-graph-convolution}}.


    \subsection{Synthetic Experiments}\label{subsec:synthetic-experiments}
    We generate two different synthetic graphs that we denote as \emph{bipartite} and \emph{tripartite}.
    Both are multipartite graphs, \ie their nodes are partitioned into $k$ sets such that there are no edges between nodes in the same partition.
    In the bipartite graph $k=2$, while in the tripartite $k=3$.
    The structure generation is completely random: give a node $v$ in a partition, we randomly select $d_v$ neighbours from the other partition.
    The procedure ensures that the average node degree in the graph is equal to the desired value.
    In our experiments, we set an average node degree of $5$.
    The total number of nodes is $1600$ for each graph, and they are divided into $k$ equally size partitions.
    For each node $v$, the output node label $y_v \in{1,\dots, k}$ is the partition to which the node belongs, and the input node feature $\bm{x}_v \in \real^f$ are sampled from a multivariate Gaussian distribution with zero mean and unitary variance, \ie $\bm{x}_v \sim \mathcal{N}(\bm{0}, I)$.
    It is important to note that the input features of a node do not provide any information about its partition.

    \begin{table}
        \centering
        \caption{Test results on synthetic data}
        \label{tab:synthetic_results}
        \begin{tabular}{lcc}
            \toprule
            & Bipartite             & Tripartite            \\ \midrule
            Lying-GCN   & $\bm{99.31 \pm 0.48}$ & $\bm{71.62 \pm 2.43}$ \\
            Lying-GCNII & $\bm{99.03 \pm 0.73}$ & $\bm{78.50 \pm 2.33}$ \\ \midrule
            GCN         & $94.19 \pm 2.49$      & $52.38 \pm 2.70$      \\
            GCNII       & $97.31 \pm 0.92$      & $48.16 \pm 3.92$      \\ \bottomrule
        \end{tabular}
    \end{table}

    For both datasets, we consider an experimental setup where we randomly split 60\% of the nodes for training, 20\% for validation, and 20\% for testing.
    To assess the model performances, we first carry out model selection by selecting the best configuration on the validation set, where accuracy is the metric of interest.
    After the best configuration is selected, we estimate the trained model's empirical risk on the test nodes.
    This process is repeated 10 times for different splits.
    The hyper-parameters validated are: the number of layers $l \in \{1,\dots,10\}$, the size of the hidden representation $d \in \{5,10,20\}$ which is the same for all layers, and the activation function $\sigma \in \{\tanh, \text{relu}\}$.
    The parameters are learned using Adam~\cite{kingma_adam_2015} optimiser with a learning rate is fixed at $0.01$; there is no weight decay nor dropout.
    For the Lying-GCNII, we also set $\alpha=0.1$ and $\lambda=1$.

    \begin{figure*}
        \centering
        \subfloat[]{
            \includegraphics[width=0.29\textwidth]{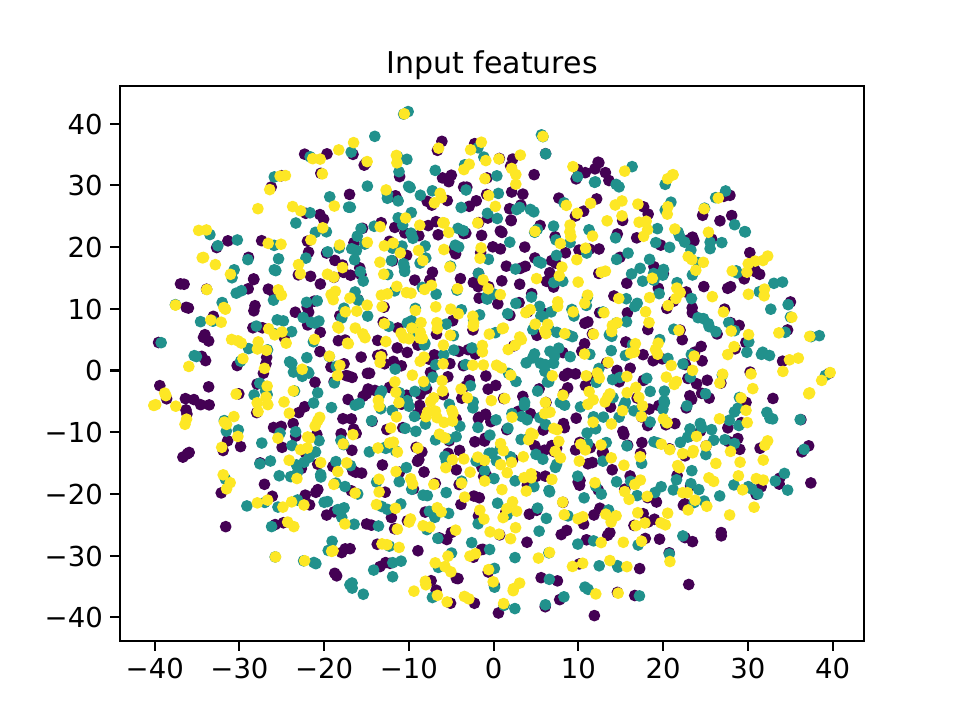}
            \label{fig:tripartite_input_features}
        }
        \subfloat[]{
            \includegraphics[width=0.29\textwidth]{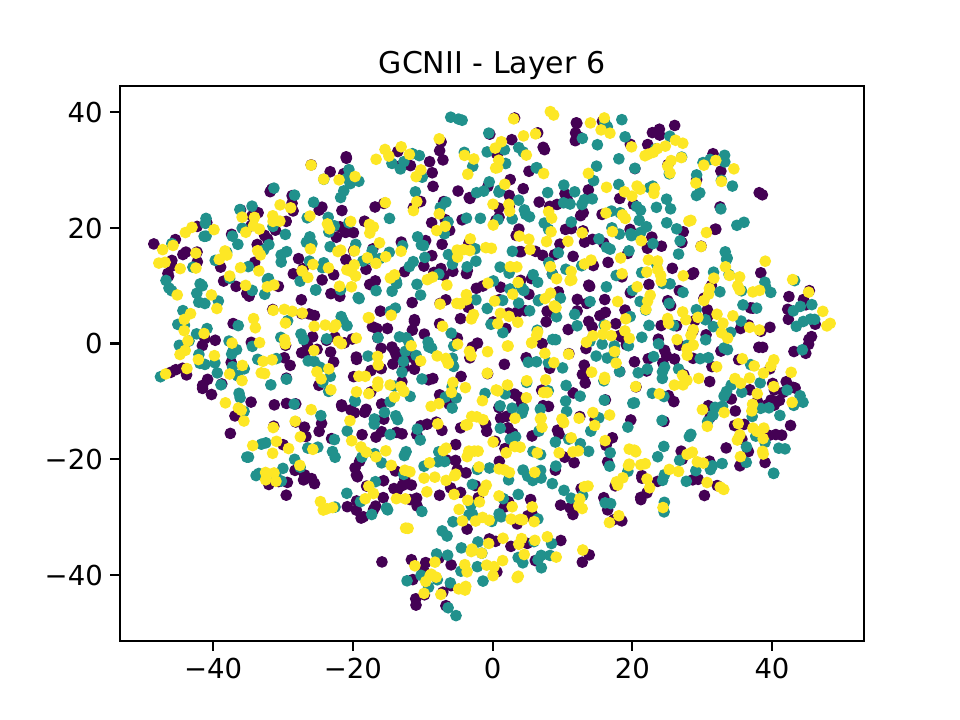}
            \label{fig:tripartite_GCNII_embs}
        }
        \subfloat[]{
            \includegraphics[width=0.29\textwidth]{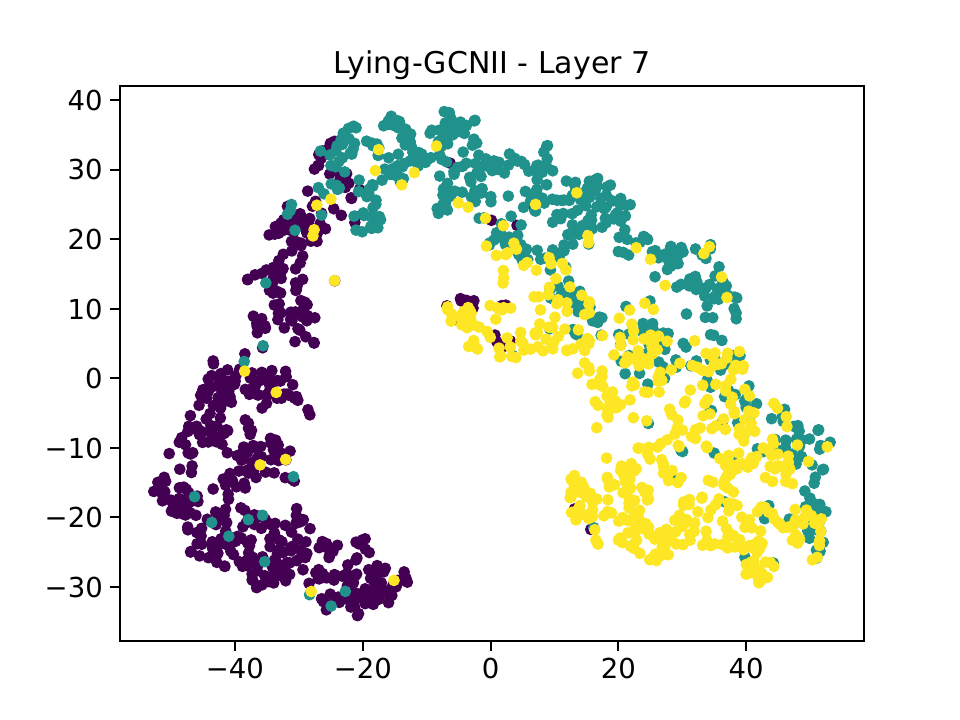}
            \label{fig:tripartite_LyingGCNII_embs}
        }
        \caption{t-SNE plot of node information on the tripartite graph: (a) the input features, (b) the node embeddings of the best GCNII configuration, and (c) the node embeddings of the best Lying-GCNII configuration.}
        \label{fig:tripartie_embs}
    \end{figure*}

    In Table~\ref{tab:synthetic_results} we report means and standard deviations of the test results.
    On the bipartite graph, all the models reach an accuracy higher than 90\%.
    The introduction of the lying mechanism is beneficial since both Lying-GCN and Lying-GCNII reach an accuracy of 99\%.
    The improvement over GCN and GCNII is marginal (5\% for the Lying-GCN and 2\% for the Lying-GCNII) but it is statistically significant ($ \text{p-value} < 0.05$).
    On the tripartite graph, the accuracy results are much lower, suggesting that the task is more difficult.
    This is in line with recent works which show that the heterophily itself is not enough to characterise the behaviour of DGNs~\cite{castellana2023investigating,ma_homophily_2022,cavallo_2_2023}.
    The bipartite graph has an easy pattern (although completely heterophilic): to determine the label (partition) of a node, it is enough to know the partition of one of its neighbours.
    In the tripartite graph, this is no longer valid since we have to exclude two partitions to infer a node label.
    In this more complex context, the benefits of the lying mechanism are more evident: while GCN and GCNII reach an accuracy near 50\%, Lying-GCN and Lying-GCNII reach an accuracy of 72\% and 78\%, respectively.
    The advantages of our proposal are more evident by observing Figure~\ref{fig:tripartie_embs}: the plot of the embeddings obtained by the best configuration of GCNII (Figure~\ref{fig:tripartite_GCNII_embs}) does not show any clusterisation, making the classification task difficult for the linear classifier;
    actually, GCNII embeddings are almost the same of the input features plotted in Figure~\ref{fig:tripartite_input_features}.
    On the contrary, the plot of the embeddings obtained by the best configuration of Lying-GCNII (Figure~\ref{fig:tripartite_LyingGCNII_embs}) shows three evident clusters (although overlapped).
    All the plots are obtained by employing t-SNE~\cite{van2008visualizing} to map the high-dimensional hidden node representations into a two-dimensional space.

    \begin{figure}
        \centering
        \subfloat[]{
            \includegraphics[width=0.45\columnwidth]{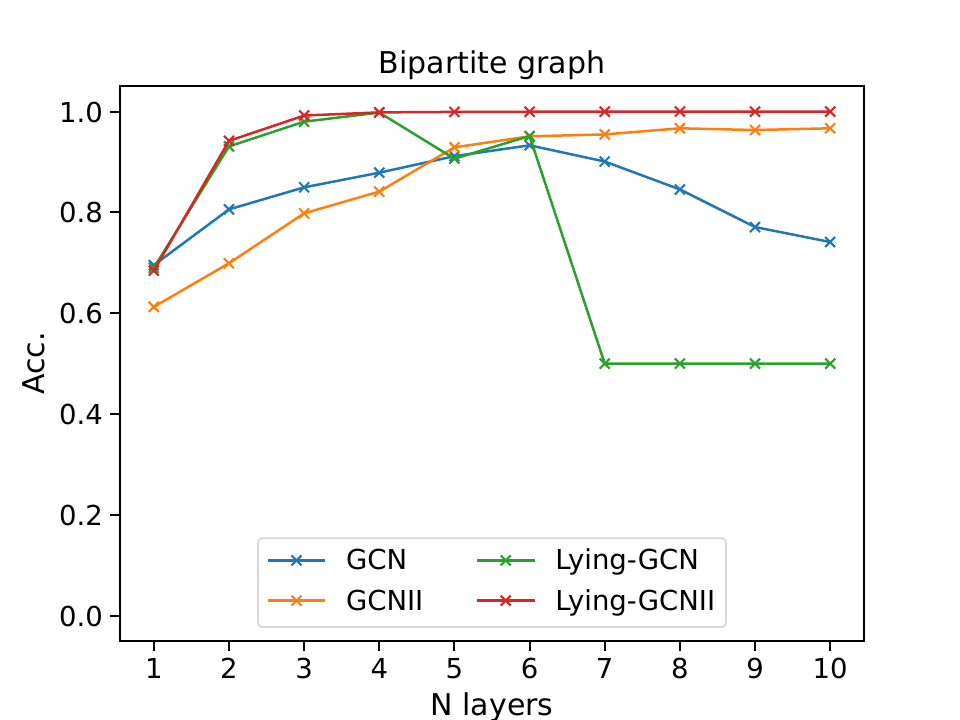}
            \label{fig:bipartite_plot}
        }
        \subfloat[]{
            \includegraphics[width=0.45\columnwidth]{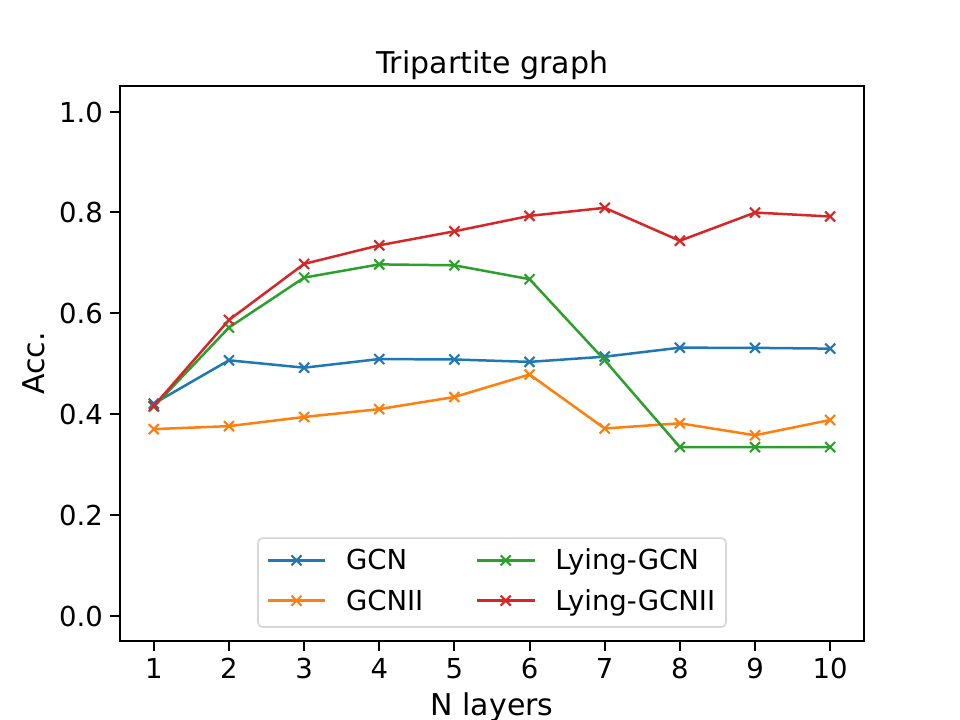}
            \label{fig:tripartite_plot}
        }
        \caption{Validation accuracy w.r.t. the number of layers for all models on synthetic datasets.}
        \label{fig:num_layers_plot}
    \end{figure}

    To better understand the behaviour of our proposed models, we plot in Figure~\ref{fig:num_layers_plot} the validation accuracy obtained by varying the number of layers.
    From the plot, it is clear that Lying-GCN suffers when the number of layers increases.
    This behaviour is not shared with Lying-GCNII, showing that the mechanisms introduced by GCNII to increase the depth of the DGN are still beneficial when applied to our proposal.
    Thus, we argue that the new propagation scheme proposed in this paper is orthogonal with respect to the mechanisms introduced by GCNII; while the former enables richer dynamics during the propagation into each layer, the latter increases the stability of deep networks obtained by stacking multiple layers.

    \subsection{Real-world Experiments}\label{subsec:real-world-experiments}
    We test our models on four real-world datasets~\cite{sen2008collective,tang2009social,pei2020geom} with different edge homophily coefficients $h$, from $h = 0.11$ (very heterophilic) to $h = 0.81$ (very homophilic).
    We do not consider other typical heterophilic datasets such as squirrel and chameleon since there is a train-test data leakage due to replicated nodes~\cite{platonov_critical_2023}.
    Also, the usage of the WebKB \footnote{\url{http://www.cs.cmu.edu/afs/cs.cmu.edu/project/theo-11/www/wwkb}} datasets introduced in~\cite{pei2020geom} is discouraged due to they very small sizes.
    Among them, we only consider Texas due to its high heterophily level.
    For all the datasets, we consider the 10 fixed splits provided in~\cite{pei2020geom}, where each split contains 48\%/32\%/20\% of nodes per class for training, validation and testing, respectively.
    To assess the model performances, we first carry out model selection by selecting the best configuration on the validation set, where again the accuracy is the metric of interest.
    After the best configuration is selected, we estimate the trained model's empirical risk on the test nodes.
    This procedure is executed independently for each split.
    The hyper-parameters validated are: the number of layers $l \in \{2,3,4,5,10,20,30\}$, the size of the hidden representation $d \in \{16,32,64\}$ which is the same for all layers, the weight decay $w_d \in {0, 0.01, 0.1}$, the probability dropout on the input feature $p_i\in \{0.4,0.6,0.95\}$, the probability dropout after each convolutional layer $p_l\in \{0.2,0.4,0.6,0.8\}$.
    For Lying-GCN, we also validated the activation function $\sigma \in \{\tanh, \text{relu}, \text{elu}\}$.
    For Lying-GCNII, the activation function $\sigma$ is the ReLU, and we validate the hyper-parameters $\alpha \in \{0.1, 0.2, 05\}$ and $\lambda \in \{0.5, 1, 1.5\}$.
    The parameters are learned using AdamW~\cite{loshchilov2018decoupled} optimiser with a learning rate is fixed at $0.01$ for both models.

    \begin{table*}
        \centering
        \caption{Test results on real-world datasets}
        \label{tab:real_world_results}
        \begin{tabular}{lcccc}
            \toprule
            & \textbf{texas} & \textbf{film} & \textbf{citeseer} & \textbf{cora} \\
            \scriptsize Hom.
            level & \scriptsize 0.11 & \scriptsize 0.22 & \scriptsize 0.74 & \scriptsize 0.81 \\
            \scriptsize \#Nodes & \scriptsize 183 & \scriptsize 7,600 & \scriptsize 3,327 & \scriptsize 2,708 \\
            \scriptsize\#Edges & \scriptsize 295 & \scriptsize 26,752 & \scriptsize 4,676 & \scriptsize 5,278 \\
            \scriptsize\#Classes & \scriptsize 5 & \scriptsize 5 & \scriptsize 7 & \scriptsize 6 \\ \midrule
            \textbf{Lying-GCN} & $72.43 \pm 7.13$ & $35.95 \pm 1.05$ & $74.84 \pm 1.44$ & $86.42 \pm 0.73$ \\
            \textbf{Lying-GCNII} & $\bm{83.51 \pm 5.73}$ & $\bm{37.05 \pm 1.02}$ & $\bm{76.33 \pm 1.86}$ & $\bm{87.78 \pm 1.43}$ \\ \midrule
            GCN & $55.14 \pm 5.16$ & $27.32 \pm 1.10$ & $\bm{76.50 \pm 1.36}$ & $86.98 \pm 1.27$ \\
            GCNII & $77.57 \pm 3.83$ & $\bm{37.44 \pm 1.30}$ & $\bm{77.33 \pm 1.48}$ & $\bm{88.37 \pm 1.25}$ \\ \midrule
            GAT & $52.16 \pm 6.63$ & $27.44 \pm 0.89$ & $\bm{76.55 \pm 1.23}$ & $86.33 \pm 0.48$ \\
            FAGCN & $\bm{82.43 \pm 6.89}$ & $34.87\pm1.25$ & N/A & N/A \\
            GGCN & $\bm{84.86 \pm 4.55}$ & $\bm{37.54 \pm 1.56}$ & $\bm{77.14 \pm 1.45}$ & $\bm{87.95 \pm 1.05}$ \\ 
            Diag-NSD & $\bm{85.67 \pm 6.95}$ & $\bm{37.79 \pm 1.01}$ & $\bm{77.14 \pm 1.85}$ & $87.14 \pm 1.06$ \\ \bottomrule
            MLP & $\bm{80.81\pm4.75}$ & $36.53\pm0.70$ & $74.02\pm1.90$ & $75.69\pm2.00$
        \end{tabular}
    \end{table*}

    In addition to GCN and GCNII, we also consider other baselines that we have already discussed in Section~\ref{sec:other_related_works}: GAT~\cite{velickovic_graph_2018}, GGCN~\cite{yan2022two}, Diag-NSD~\cite{bodnar_neural_2022} and FAGCN~\cite{bo2021beyond}.
    All the baseline results are taken from~\cite{bodnar_neural_2022} and they are obtained on the same set of splits as ours.
    All the results are reported in Table~\ref{tab:real_world_results}, where results in bold are not statistically different ($ \text{p-value} > 0.05$) w.r.t the state of the art.
    The results confirm what we have observed on the synthetic datasets: the lying mechanism is useful when the dataset is heterophilic (Lying-GCN and Lying-GCNII outperform GCN and GCNII by 17\% and 6\% on texas, respectively).
    If we focus on homophilic datasets (citeseer and cora), the results of our proposal are comparable with GCN and GCNII: in particular, Lying-GCNII obtain results that are statically not different from the state-of-art.
    This suggests that our approach is not harmful and that it adapts to the characteristics of the input graph.
    Also, Lying-GCNII always outperforms Lying-GCN highlighting the benefit of a more stable and deep architecture.
    Among the baseline models (the bottom part of the table), GAT is the only model that cannot set negative weight over edges: on all the datasets except citeseer, Lying-GCNII outperforms GAT. The huge difference in performances on texas and film datasets highlights the benefits of negative edge weights in the heterophilic setting.
    On the film dataset, Lying-GCNII outperforms FAGCN even if the two approaches are similar: this highlights the benefit of learning a different edge weight for each channel.
    GGCN and Diag-NSD are very challenging baselines based on negative edge weights, and our proposal Lying-GCNII obtains comparable results on all the datasets.
    Also, it is worth pointing out that both GGCN and Diag-NSD employ other mechanisms that could be helpful in practice: for example, GGCN learns a set of weight for each layer to combine the self-representation of a node ($\bm{h}_v$) with the sum of the message received;
    similarly, Diag-NSD augments the restriction maps with fixed values (1 and -1) to improve the diffusion when the node representations are not good enough to learn good maps.
    Both strategies could be integrated with the lying diffusion process.

    \section{Conclusion} \label{sec:conclusion}
    In this paper, we introduce Lying-GCN, a new DGN based on opinion dynamics.
    The key of our proposal is the introduction of a lying mechanism which allows nodes to not share their hidden representation directly.
    Instead, each node shares different messages to each neighbour by multiplying its hidden representation with a set of weights obtained through an adaptive procedure.

    We also provide a characterisation of the lying mechanism in terms of dynamical systems.
    On the one hand, our theoretical results show that the system asymptotically collapses to zero, meaning that the DGN is prone to oversmoothing.
    On the other hand, they highlight an oscillatory pattern in the early stages of the diffusion due to the presence of complex eigenvalues.

    The empirical results confirm that Lying-GCN improves the performances of GCN in the heterophilic setting without harming in the homophilic one, notwithstanding its performance degradation when the number of layers increases.
    Interestingly, the poor performances obtained by deep networks can be counteracted by employing the two techniques introduced by GCNII. On four real-world datasets, Lying-GCNII is able to perform on par with state-of-the-art models.
    Thus, we argue that the new propagation scheme proposed in this paper is orthogonal to the mechanisms introduced by GCNII; while the former enables richer dynamics during propagation, the latter increases the stability of deep networks obtained by stacking multiple layers.

    Our findings highlight that the interplay between performances in the heterophilic setting and oversmoothing phenomena needs to deepen.
    In fact, while recent papers such as~\cite{bodnar_neural_2022, yan2022two} show that the two phenomena have common causes, we show that they can be also addressed separately.
    In particular, it is interesting that the lying mechanism that we introduce to increase the performances on heterophily graphs operates on the propagation schema (\ie it modifies the Laplacian), while the techniques which reduce the oversmoothing (\eg~\cite{chen2020simple,gravina2023antisymmetric}) acts on the weight matrices.
    To this end, it would be interesting to develop a new adaptive propagation schema that can guarantee also the absence of oversmoothing.
    We would like to develop and study such a schema through the lens of tensor theory, which has been proven to be effective on structured data~\cite{castellana2020coling, castellana2021neurocomp, hua2022high}.

    \bibliographystyle{IEEEtran}
    \bibliography{main}

    \appendix
    \label{sec:appendix}
    \begin{IEEEproof}
        [Proof of Proposition~\ref{prop:real_eigenvalue_E}]
        First of all, we observe that the Laplacian matrix $\tilde{L} = \tilde{D} - \tilde{A} = D-A$ is diagonally dominant since $\tilde{L}_{uu} = g_u = \sum_{v\neq u} |\tilde{L}_{uv}|$, where $g_u$ is the degree of node $u$.
        By the definition of $Z$, it holds that $Z_{uu} = 0$ since we assume there are no self-loops in the original input graph and $Z_{uv} \in [-1,1]$ since it is the image of the $\tanh$ function.
        It follows that also the matrix $B=\tilde{L} \odot (Z+I)$ is diagonally dominant since $B_{uu} = \tilde{L}_{uu}=g_u$ and $\sum_{v\neq u} |B_{uv}| = \sum_{v\neq u} |\tilde{L}_{uv}|\cdot|Z_{uv}|\leq \sum_{v\neq u} |\tilde{L}_{uv}| = g_u$.
        By Gersgorin’s theorem~\cite{varga2010gervsgorin}, the real part of each nonzero eigenvalue of $B$ is strictly positive.
        Finally, we conclude by showing that the eigenvalues of $E$ and $B$ have the same sign since $E = \tilde{D}^{-\frac{1}{2}}B\tilde{D}^{-\frac{1}{2}}$ and $\tilde{D}^{-\frac{1}{2}}$ has strictly positive entries.
    \end{IEEEproof}
\end{document}